\documentclass[a4paper,fleqn]{cas-sc}
\usepackage{subfigure}
\usepackage{graphicx}
\usepackage{float}
\usepackage{threeparttable}
\usepackage{hyperref }
\usepackage{url}
\usepackage{diagbox}
\usepackage{ulem}
\usepackage{caption}
\usepackage{algorithm}%
\usepackage{algpseudocode}%
\usepackage{lineno} 
\usepackage[authoryear]{natbib}
\usepackage{float}
\def\tsc#1{\csdef{#1}{\textsc{\lowercase{#1}}\xspace}}
\tsc{WGM}
\tsc{QE}
\tsc{EP}
\tsc{PMS}
\tsc{BEC}
\tsc{DE}
\begin{document}
% \pagewiselinenumbers% 按页重新编号 
% \linenumbers
\begin{sloppypar}
\let\WriteBookmarks\relax
\def\floatpagepagefraction{1}
\def\textpagefraction{.001}
\shorttitle{Information Fusion}
\shortauthors{Hao Lyu et~al.}
%\begin{frontmatter}

\title [mode = title]{Knowledge-data fusion dominated vehicle platoon dynamics modeling and analysis: A physics-encoded deep learning approach}                      
\author[1,2,3]{Hao Lyu}[orcid=0000-0002-1664-8050]
\ead{lyu_hao@seu.edu.cn}

\address[1]{School of Transportation, Southeast University, Nanjing, China, 211189}
\address[2]{Jiangsu Key Laboratory of Urban ITS, Nanjing, China, 210096}
\address[3]{Jiangsu Collaborative Innovation Center of Modern Urban Traffic Technologies, Nanjing, China, 210096}
\address[4]{Department of Automation, Beijing National Research Center for Information Science and Technology (BNRist), Tsinghua University, Beijing, China, 100084}

\author[1,2,3]{Yanyong Guo}
\cormark[1]
\ead{guoyanyong@seu.edu.cn}
 
\author[1,2,3]{Pan Liu}
\ead{lyuhao@seu.edu.cn}

\author[4]{Shuo Feng}
\ead{fshuo@tsinghua.edu.cn}

\author[1,2,3]{Weilin Ren}
\ead{230228916@seu.edu.cn}

\author[1,2,3]{Quansheng Yue}
\ead{yue_qs@seu.edu.cn}

\begin{abstract}
Recently, artificial intelligence (AI)-enabled nonlinear vehicle platoon dynamics modeling plays a crucial role in predicting and optimizing the interactions between vehicles. Existing efforts lack the extraction and capture of vehicle behavior interaction features at the platoon scale. More importantly, maintaining high modeling accuracy without losing physical analyzability remains to be solved. To this end, this paper proposes a novel physics-encoded deep learning network, named PeMTFLN, to model the nonlinear vehicle platoon dynamics. Specifically, an analyzable parameters encoded computational graph (APeCG) is designed to guide the platoon to respond to the driving behavior of the lead vehicle while ensuring local stability. Besides, a multi-scale trajectory feature learning network (MTFLN) is constructed to capture platoon following patterns and infer the physical parameters required for APeCG from trajectory data. The human-driven vehicle trajectory datasets (HIGHSIM) were used to train the proposed PeMTFLN. The trajectories prediction experiments show that PeMTFLN exhibits superior compared to the baseline models in terms of predictive accuracy in speed and gap. The stability analysis result shows that the physical parameters in APeCG is able to reproduce the platoon stability in real-world condition. In simulation experiments, PeMTFLN performs low inference error in platoon trajectories generation. Moreover, PeMTFLN also accurately reproduces ground-truth safety statistics. The code of proposed PeMTFLN is open source\footnote{\url{https://github.com/SpaceTrafficSafetyTeam/PeMTFLN}}.
\end{abstract}


\maketitle

\section{Introduction}
Accurate vehicle dynamic modeling provides theoretical support for understanding the microscopic behavior between vehicles. More importantly, it is the foundation of critical applications in intelligent transportation systems such as traffic flow simulation, platoon formation control, and signal optimization \citep{
li2023modified, qin2022demand}. Car following (CF) behavior that reflects the coupling mechanism between the current and preceding vehicles on the same lane is the most important component of dynamic modeling. Starting from pioneering works such as the Pipes model \citep{pipes1953operational} and the optimal velocity (OV) model \citep{bando1995dynamical}, car following models have been attracted in the field of traffic engineering and statistical physics. The purpose is to reveal nonlinear characteristics such as traffic instability, shock waves, and phase transitions, to elucidate the mechanism of traffic congestion formation and dissipation. However, the trajectories generated by physics-driven models based on ideal assumptions are naturally more conservative, leading to deviations in replicating human driving behavior. In recent years, disruptive deep learning has shined in the application of intelligent transportation systems \citep{wang2024pi,lu2025hyper}. Focusing on car-following modeling, coupled with the emergence of high-precision, large-sample vehicle trajectory data, deep learning has also spurred the emergence of data-driven car following models \citep{zhou2023data}. Generally, the goals of mathematical elegance and data accuracy are pursued in data-driven models. Currently, a novel physics-informed deep learning (PIDL) framework has emerged and received increasing attention \citep{karniadakis2021physics, wang2024knowledge}. However, existing PIDL-based CF models respect the physical laws by incorporating a weakly imposed loss function consisting of the physics equations residual and boundary constraints. These models are limited to theoretical considerations and implementation application considerations.

Moreover, the propagation of traffic disturbances is more significant in vehicle platoon \citep{ngoduy2021noise,li2011characterization}, and it is necessary to focus on platoon dynamics modeling. Looking further ahead, platoon dynamics modeling also caters to the needs of autonomous driving platoon simulation and testing \citep{feng2023dense,zhu2022flow}. Research efforts around platoon-level car following dynamics modeling are primarily realized through the distributed extension of vehicle-level car following modeling \citep{xu2024sequence}. Such direct extension methods will amplify modeling inaccuracies and ignore complex vehicle interactions within the platoon. What’s more, the interaction between vehicles is dynamically evolving and exhibits temporal correlation. Even for the mixed vehicle platoon, previous studies often relax the assumption of homogeneous vehicles and do not fully focus on the overall characteristics of the platoon \citep{yu2024theory}. It is a critical need to learn and explore the intrinsic overall characteristics of car following behavior from the perspective of the overall platoon. 

To address the above limitations, this paper proposes a physics-encoded deep learning framework for vehicle platoon dynamics modeling and analysis, named PeDL. The core modeling perspective is to focus on the power of parameters of the generalized car following model and forcibly encode known physics into the core architecture. Specially, we designed an analyzable parameters encoder computational graph (APeCG) and a multi-scale trajectory feature learning network (MTFLN) within the PeDL framework. APeCG is oriented by the generalized car following model, with the aim of directly learns analyzable parameters. MTFLN enables the extraction of interaction features of vehicle-level and platoon-level, covering both intrinsic and dynamic features at different scales. Notably, the training of PeDL is end-to-end. The main contributions are as follows:

\begin{itemize}
\item  A novel vehicle platoons dynamics modeling framework (PeDL) with interpretability and accuracy is proposed. It focuses on learning and encoding the parameters of the generalized platoon model, aiming to avoid excessive dependence on prior knowledge, and to maximize the utility of the framework. Moreover, PeDL can be scaled to model platoons with different numbers of vehicles and highly interactive scenarios.

\item A multi-scale feature learning network (MTFLN) is designed to facilitate the end-to-end learning of the parameters necessary for PeDL. The network initially extracts vehicle-level features from naturalistic driving platoon trajectories. With these features as the center, the causal attention mechanism is employed to learn comprehensive platoon-level features and implements multi-step prediction. 

\item A platoon trajectory dataset is extracted from HIGH-SIM dataset \citep{shi2021video} and applied to PeDL. The proposed PeDL can reproduce the ground-truth of platoon following behavior with high accuracy. In particular, PeDL succeeded in replicating both the stability and safety evolution of the platoon, highlighting its exceptional capacity in physical analyzability.
\end{itemize}

The remainder of the paper is structured as follows. Section \ref{2} reviews related work on vehicle dynamic modeling at the vehicle-level and platoon-level. In Section \ref{3}, we describe the problem and introduce the proposed PeMTFLN. Section \ref{prediction} adopts real-world trajectories data to train and evaluate the prediction performance of the PeMTFLN. Section \ref{Simulation} presents simulation experiments and results analysis on platoon trajectories generation. The last Section \ref{Conclusion} concludes the paper and discusses follow-up work.

\section{Related Works\label{2}}
\subsection{Vehicle-level dynamics modeling}
%physics
The physics-driven car following model is defined as a rigorous derivation process, with most parameters having clear physical meanings. Classic car following models such as full velocity difference model (FVD) \citep{zhao2005new} and intelligent driver model (IDM) \citep{treiber2000congested}, greatly promoted the fundamental research of car following behavior. Subsequently, there are various perspectives to model car following behavior, such as stochasticity \citep{ngoduy2019langevin} and heterogeneity \citep{wang2022stability}. 
%Data
In terms of data-driven models, Ma et al. considered reaction delay and established a sequence to sequence learning based car-following model \citep{ma2020sequence}. Liao et al. develops a personalized car-following model using a memory-based deep reinforcement learning approach, specifically integrating LSTM with twin delayed deep deterministic policy gradients, to optimize driving behavior \citep{liao2024modelling}. Jiang et al. presented a stochastic learning approach using approximate Bayesian computation to integrate multiple car-following models, achieving more accurate vehicle trajectory reproduction for both human-driven and automated vehicles \citep{jiang2024generic}. 
%PIDL
Innovatively, Mo et al. introduced a physics-informed deep learning encoded with physics-based models, providing a new perspective and tool for car following research \citep{mo2021physics}. Afterwards, Transformer neural network architecture with self-attention was applied to the PIDL-based car following architecture, achieving superior performance \citep{geng2023physics}. However, within the PIDL-based car following models, physical prior knowledge only played a secondary role in training deep neural networks, where the neural network remains the main focus of inference. An additional drawback is that these models idealize assumptions for modeling actual traffic flow and lack of the analysis of traffic characteristics of real traffic flow.

\subsection{Platoon-level dynamics modeling}
While most research has focused on predicting short-term trajectories of individual vehicles, there is a growing need to model platoon-level car-following dynamics to better understand the interactions and collective behavior of multiple vehicles. For example, Lin et al. analyzed the spatiotemporal error propagation from the perspective of platoon generation, and proposed an long short-term memory network structure with improved sampling mechanism \citep{lin2020platoon}. Lin et al. worked on providing enhanced state representation for multi-agents reinforcement learning, incorporating intricate relative vehicle relationships to improve performance in platoon-following scenarios \citep{lin2024enhancing}. Liu et al. combined physics knowledge with deep learning and stochasticity to design a quantile-regression physics-informed deep learning car-following model, which comprehensively characterizes the real laws of car following behavior in a platoon \citep{liu2023quantile}. Furthermore, Rowan et al. attempted to capture mesoscopic and macroscopic traffic phenomena as broader traffic features for model training \citep{davies2024multiscale}, and constructed a multiscale car-following framework for vehicle platoons. Tian et al. proposed an AI-based Koopman framework to model the unknown nonlinear platoon dynamics. This framework linearizes the platoon system, aiming to analyze how disturbances propagates over vehicle platoon while achieving accurate prediction \citep{tian2024physically}. However, due to the existence of operators, this framework lacks scalability. Overall, there is a lack of platoon-level dynamic modeling, particularly in capturing the behavioral interaction features across the physical space of vehicles at the platoon-level. More importantly, the existing platoon-level dynamics modeling lacks of addressing the scalability and analyzability, issues simultaneously, which motivates the focus of this paper.

\section{methodology\label{3}}
\subsection{Problem Statement}
A vehicle platoon traveling in the same lane is under investigation. This platoon comprises a leading vehicle, designated as vehicle 0, and a set of following vehicles, numbered from 1 to $N$. The problem of modeling the dynamics of the platoon is defined as follows: given the historical state sequences of the $N$ following vehicles, the objective is to model the platoon dynamics and predict the state sequences of the following vehicles over a specified prediction horizon. This prediction is based on the future velocity sequences of the leading vehicle. To imbue the model with physical interpretability and the ability to provide closed-loop responses, the problem can be formalized as follows:
\begin{equation}
    \label{eq1}
    \begin{array}{l}
    \theta_t \leftarrow {g}\left( {{X_{1:N}}\left( {t - P:t} \right)} \right)\\
    {x_{1:N}}\left( {t+1:t + F} \right) =  f_{\theta_t}\left( {{x_{1:N}}\left( t \right),{v_0}\left( {t+1:t + F} \right)} \right)
    \end{array}
    \end{equation}
where $P$ and $F$ denote the input historical horizon and the output prediction horizon, respectively; $x_n(t) = \left[ v_n(t), s_n(t), \Delta v_n(t) \right]^T$ denotes the state of vehicle $n$ at time $t$. This state vector includes the velocity of vehicle $n$, $v_n(t)$, the gap to the preceding vehicle, $s_n(t)$, and the relative velocity to the preceding vehicle, $\Delta v_n(t) =v_{n-1}(t) -  v_n(t)$. Additionally, $X_n(t)\in {{R}^{ d_{in}}}$ represents the extended state vector, which encompasses additional information such as vehicle type, etc, $d_{in}$ represents the dimension of $X_n(t)$. The notation $X_{1:N}(t - P:t)\in {{R}^{P\times N \times d_{in}}}$ refers to the state sequences of vehicles 1 to $N$ over the interval from time $t - P$ to $t$; $\theta_t$ symbolizes the parameters of the platoon dynamics model, $g(\cdot)$ denotes the mapping from the state sequences to these parameters, and $f_{\theta_t}(\cdot)$ represents the platoon dynamics model parameterized by $\theta_t$. The problem of modeling the platoon is redefined as 1) reasonably parameterizing $f(\cdot)$ and identifying the necessary modeling parameters $\theta_t$; and 2) developing the mapping $g(\cdot)$ to optimize parameters, enabling the $f_{\theta_t}(\cdot)$  model to generate human-like trajectories of vehicle platoons based on real-time variations in the velocity of the leading vehicle.

\subsection{Physics-Encoded Deep Learning Framework}
The platoon dynamics model should be capable of modeling platoons with varying numbers of vehicles to accommodate different traffic scenarios. Additionally, the extensive body of knowledge on vehicle-level car-following behavior provides a critical foundation for modeling platoon-level dynamics. Consequently, the formulation of $f(\cdot)$ should adhere to the established principles of car-following behavior. The generalized formulation of the car-following model given is as follows:
\begin{equation}
    \label{eq2}
   {a_{n}}\left( {t} \right) =  f_{n}\left( {{x_{n}}\left( t \right)} \right)
 \end{equation}
In theory-based models, $f_{n}(\cdot)$ is typically characterized by parameters that are physically interpretable. These parameters often provide a quantitative description of driver characteristics, enabling $f_{n}(\cdot)$ to describe and explain a driver’s decision-making based on the current state. However, many of these models are accompanied by idealized assumptions, thereby limiting their ability to capture diverse driving behaviors effectively. 
The proposed framework employs a linear time-varying form of $f_{n}(\cdot)$, thereby avoiding the need for specific model choices and excessive reliance on prior knowledge. The linear  time-varying formulation of Eq. (\ref{eq2}) is presented as : 
\begin{equation}
    \label{eq3}
    \begin{array}{c}
  {a_{n}}\left( {t} \right)    \approx {f_{n,t}}(x_n^ * (t)) + {f_{x_n^ * (t)}}^\prime  \cdot ({x_n}(t) - x_n^ * (t))
  = {f_{x_n^ * (t)}}^\prime  \cdot ({x_n}(t) - x_n^ * (t))
  \end{array}
 \end{equation}
where $x_n^ * (t)$ denotes the expected state vector of vehicle $n$ at time $t$; ${f_{x_n^ * (t)}}^\prime  = \left[ {{f_{v_n^ * (t)}}^\prime ,{f_{s_n^ * (t)}}^\prime ,{f_{\Delta v_n^ * (t)}}^\prime } \right]$ represents the partial derivative  vector of $f_{n}(\cdot)$ with respect to the expected states. Their values indicate the degree to which drivers aim to achieve or maintain the corresponding states as desired values. Furthermore, the physical meaningfulness of $f_{n}(\cdot)$ necessitates the following conditions: 
\begin{equation}
    \label{eq4}
  {{f_{v_n^ *}}^\prime<0 ,{f_{s_n^ * }}^\prime>0 ,{f_{\Delta v_n^ *}}^\prime>0 } 
 \end{equation}
Eq. (\ref{eq4}) carries intuitive physical implications. For instance, regarding velocity, when a vehicle's actual velocity surpasses the expected velocity, drivers tend to decelerate, and conversely so. Moreover,  Eq. (\ref{eq4}) represents a sufficient condition for ensuring local stability within the model. Furthermore, Eq. (\ref{eq4}) is parameterized to replicate and expound upon specific driving behaviors:
\begin{equation}
\begin{array}{l}
    \label{eq5}
 \left[ {{f_{v_n^ * (t)}}^\prime ,{f_{s_n^ * (t)}}^\prime ,{f_{\Delta v_n^ * (t)}}^\prime } \right] \buildrel \Delta \over = {\theta _{n,t}}\in {{R}^3}
 \quad x_n^ * \left( t \right) \buildrel \Delta \over = {\left[ {mean({v_n}(t - P,t),mean({s_n}(t - P,t),0} \right]^T}
  \end{array}
 \end{equation}
where $\theta _{n,t}$ denotes the time-varying parameters to be determined. 

The expected states of velocity and gap are defined as the average states of the vehicle over the past $P$ timestamps. This definition rests on a reasonable assumption: drivers consistently desire vehicles to maintain a velocity that aligns with the preceding vehicle's velocity, while minimizing rapid fluctuations in speed and gap over the short term. Eqs. (\ref{eq2})-(\ref{eq5}) were extended spatially and temporally to encompass the $N$ following vehicles within the platoon and across the prediction horizon and obtained the form of parameters $\theta_t\in {{R}^{N\times F \times3}}$ that need to be determined for platoon dynamic modeling. However, solving for $\theta_t$ remains inherently difficult. The proposed framework employs advanced neural networks to explore the mapping $g(\cdot)$, which relates naturalistic driving platoon trajectories to the parameter $\theta_t$.

\textbf{Remark 1}: Linear time-varying model ensures a balanced trade-off between computational complexity and modeling accuracy. The influence of platoon-level characteristics on individual vehicle driving behavior is reflected through implicit adjustments in parameters $\theta_t$.
\subsection{Model Architecture}
\begin{figure}
  \vspace{-1em}  % 调整与上文的间距
    \centering
  \includegraphics[width=0.9\textwidth]{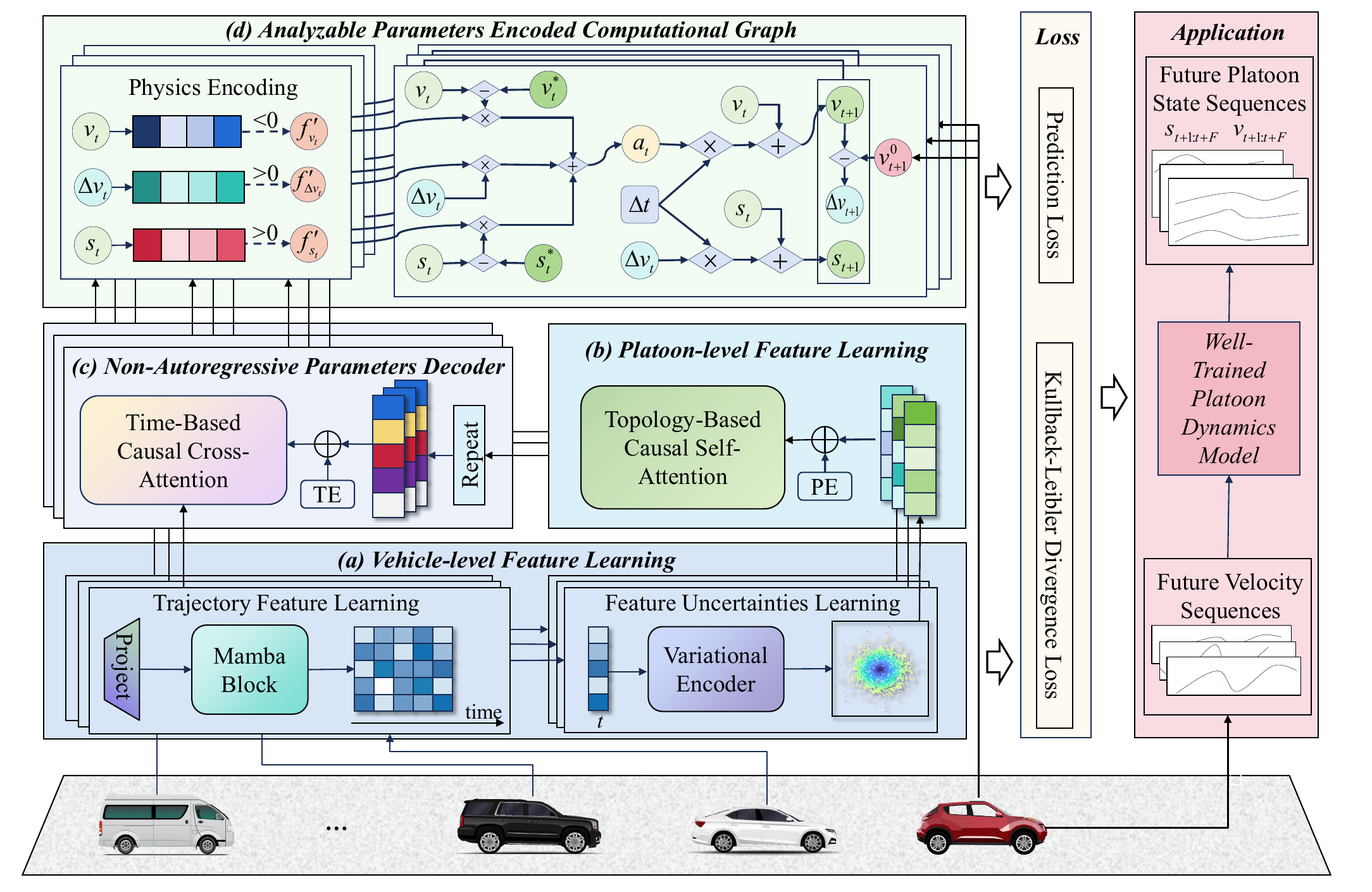}
  \caption{The architecture of PeMTFLN under PeDL Framework}\label{total}
  \vspace{-1em}  % 调整与上文的间距
\end{figure}

The architecture of parameters encoder multi-scale trajectory feature learning network (PeMTFLN) is illustrated in Fig. \ref{total}. MTFLN is depicted by components (a) to (c), whereas APeCG is represented by component (d) in Fig. \ref{total}. Specifically, the historical trajectories of the following vehicles within the platoon are input into the Vehicle-level Feature Learning (VFL) module. The VFL module comprises two sub-modules: Trajectory Feature Learning (TFL) and Feature Uncertainty Learning (FUL). The first sub-module extracts relevant features from the vehicle's historical trajectories, while the second generates probabilistic representations of these features. The VFL module is applied in parallel to each vehicle, with shared weights across all instances, ensuring the model's scalability. Following this, vehicle driving characteristics samples, randomly drawn from the distribution output by the FUL sub-module, are used as tokens and fed into the Platoon-level Feature Learning (PFL) module to learn interactions among vehicles within the platoon. The output of the PFL module serves as the query, while the outputs of the TFL sub-module as the keys and values, which are then input into the Non-Autoregressive Parameters Decoder (NARP Decoder) to predict the value of the necessary parameters. APeCG module initially assigns physical meaning to the outputs of decoder according to Eq.(\ref{eq4}). It then calculates the platoon state over the prediction horizon based on these parameters.

\subsection{Model Details}
% This section provides a detailed introduction to the design of each module in the proposed PeMTFLN.
\subsubsection{Vehicle-level Feature Learning Module}
The VFL module is dedicated to extracting key features from a single vehicle's historical trajectories that are pertinent to driver behavior and decision-making. This module is divided into the following two components.

\begin{figure}
  \centering
  \vspace{-1em}  % 调整与上文的间距
  \includegraphics[width=0.8\textwidth]{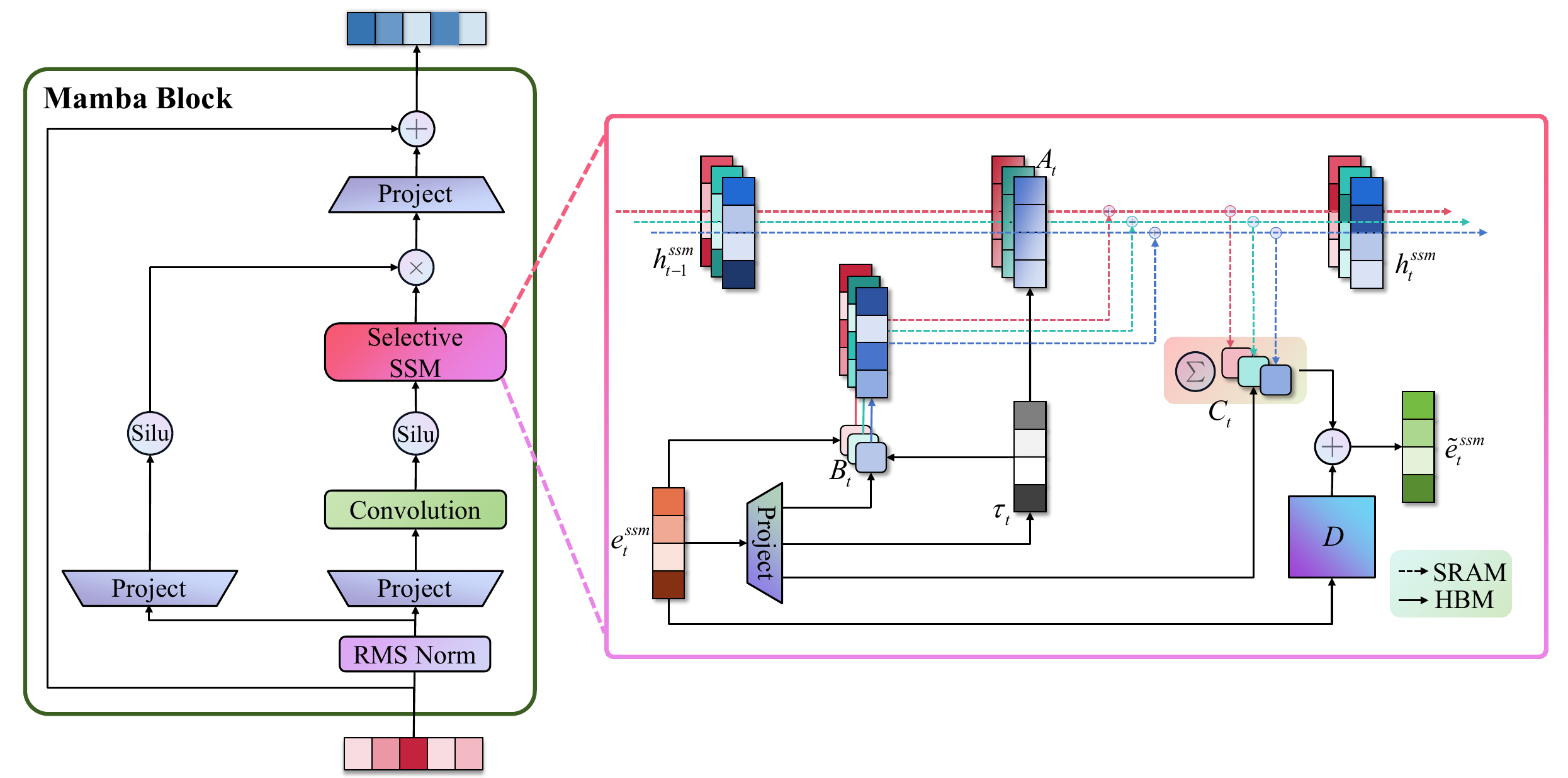}
  \caption{The structure of the mamba block in TFL module}\label{mamba}
  \vspace{-1em}  % 调整与上文的间距
\end{figure}

\textbf{1) Trajectory Feature Learning. }The continuous-time vehicle state sequence encapsulates information at various levels, including state transition relationships at different scales, the inherent and dynamic features of vehicles underlying these relationships. Within the TFL sub-module, the Mamba Block \citep{gu2023mamba,shao2025st} is employed to learn and extract these multidimensional features. The structure of the Mamba Block is depicted in Fig. \ref{mamba}. After essential normalization and convolution operations for feature enhancement, the input is fed into the state-space model (SSM) as shown in the right of Fig. \ref{mamba}. This SSM is designed to simulate vehicle state changes through state transition processes across multiple dynamic time scales. In the right of Fig. \ref{mamba}, each colored line represents a subsystem. The system matrices of these subsystems are learned across the entire dataset to capture the common features of vehicles, forming the basis of system evolution. Other parts of the subsystems dynamically adapt based on the input. For example, at time $t$, the SSM input $e_t^{ssm}$ is used to derive the control vector $B_t$ and the observation weight vector $C_t$ for each subsystem. This enhances the model's capability to perceive dynamic inputs and generalize effectively. All subsystems share the same time step $\tau_t $, and the $\tau_t $ corresponding to each dimension is different. This enhances the model's stability. Furthermore, Mamba achieves an optimal balance of performance and efficiency through selective scan and hardware-aware algorithm optimization. Overall, the TFL sub-module, with Mamba as its core, is specifically designed for vehicle trajectory feature extraction, and its unidirectional extraction of temporal information obviates the need for input position encoding. The computation formula for the TFL module is as follows:
\begin{equation}
    \label{eq6}
 H_{n,t - P:t}^{TFL} = {\varPhi  _{TFL}}({X_n}(t - P:t),{W_{TFL}})\quad n = 1,2, \cdots ,N
 \end{equation}
where $H_{n,t - P:t}^{TFL}\in {{R}^{P \times d_{m}}}$ denotes the output of the TFL sub-module,  $W_{TFL}$ and $\varPhi _{TFL}$ represent the learnable parameters and operations within the TFL sub-module, respectively. $d_{m}$ signifies the dimensionality of the model.

\textbf{2) Feature Uncertainty Learning.} Human drivers often exhibit inherent randomness in their driving behavior. In other words, even if two vehicles possess identical attributes and historical trajectories, their future driving behaviors cannot be guaranteed to be consistent. To model this randomness, the FUL sub-module incorporates a variational encoder (VE) \citep{kang2023trajectory}. Specifically, the output of the TFL sub-module $h_{n,t}^{TFL}$, which encapsulates the driver's characteristics and the vehicle state at time $t$, is fed into the VE. Leveraging knowledge acquired from extensive driving trajectories, the VE models the latent distribution of driving features for this input and outputs a parameterized driving feature distribution. The VE is implemented using a multilayer perceptron (MLP). The formal expression of the FUL sub-module is as follows:
\begin{equation}
    \label{eq7}
 \begin{array}{*{20}{l}}
{{\mu _{n,t}},\sigma _{n,t}^2 = {\varPhi _{FUL}}(h_{n,t}^{TFL},{W_{FUL}})}\\[1ex]
{{P_{W_{FUL}} }\left( {h_{n,t}^{FUL}\left| {h_{n,t}^{TFL}} \right.} \right)\sim {\cal N}\left( {{\mu _{n,t}},{\Sigma _{n,t}}} \right)}
\end{array}\; n = 1,2, \cdots ,N
 \end{equation}
where $\mu _{n,t}\in {{R}^{d_{m}}}$ and $\sigma _{n,t}^2\in {{R}^{d_{m}}}$ are the outputs of the VE. The driving feature representation distribution $P_{W_{FUL}}$ for vehicle $n$ at time $t$ is defined as an independent multivariate Gaussian distribution with mean $\mu _{n,t}$ and variance $\sigma _{n,t}^2$. $W_{FUL}$ represents all the parameters required for variational inference. $\Sigma_{n,t} $ denotes a diagonal matrix with $\sigma _{n,t}^2$ as its diagonal elements. $h_{n,t}^{FUL}\in {{R}^{d_{m}}}$denotes a sample drawn from distribution $P_{W_{FUL}}$.

The VFL module is scalable to platoons of any length. Its output includes the historical trajectory features $H_{1:N,t - P:t}^{TFL}\in {{R}^{N \times P \times d_{m}}}$ and sample $H_{1:N,t}^{FUL}\in {{R}^{N \times d_{m}}}$ of the driving features for each vehicle in the platoon.

\subsubsection{Platoon-level Feature Learning Module}
\begin{figure}
  \centering
  \vspace{-1em}  % 调整与上文的间距
  \includegraphics[width=0.8\textwidth]{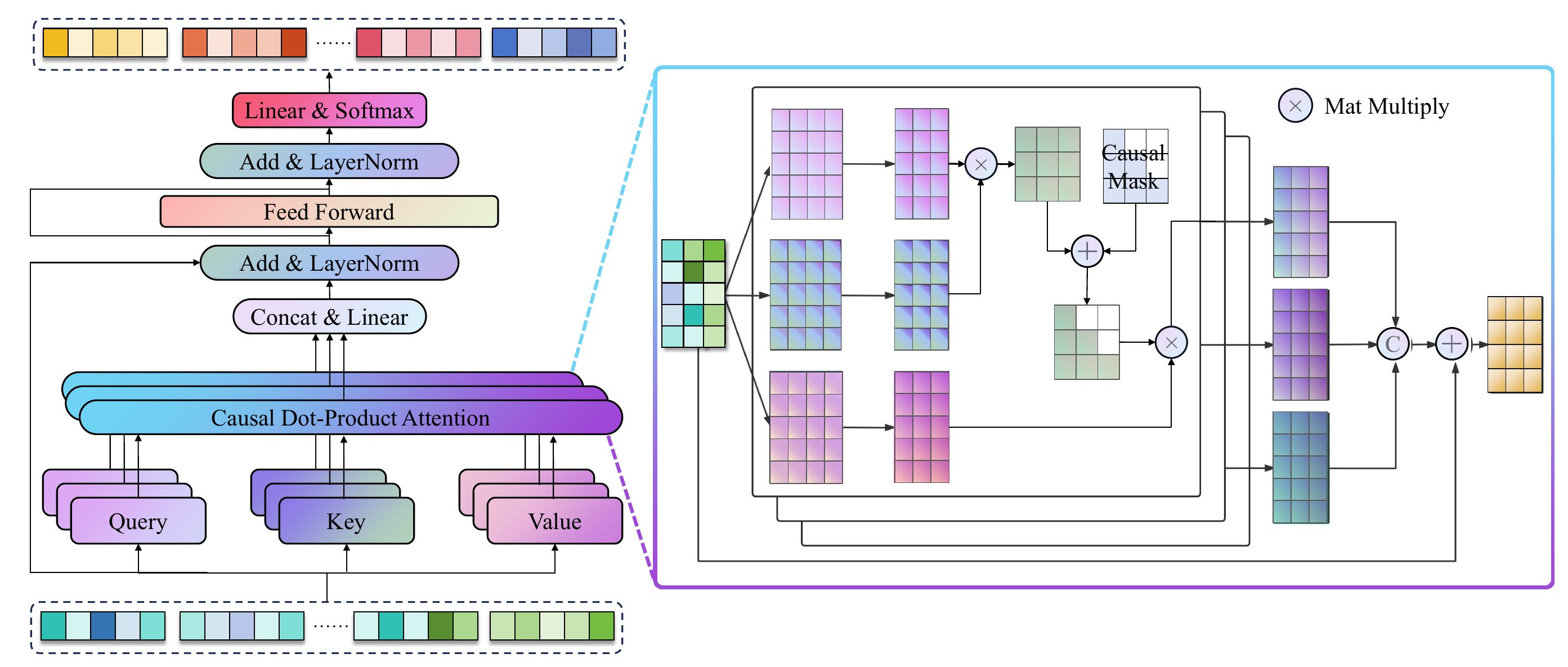}
  \caption{The structure of the topology-based causal attention in PFL module}\label{feature}
  \vspace{-1em}  % 调整与上文的间距
\end{figure}

Although HDVs cannot observe the states of multiple preceding vehicles, small changes of state of the leading vehicle can be progressively amplified in closely driving platoons, significantly impacting the following vehicles in the platoon. This often compels drivers to alter their driving decisions. As illustrated in Fig. \ref{feature}, the PFL module leverages causality-based self-attention \citep{ge2023causal} grounded in platoon topology to learn platoon-level interaction features. The attention mechanism's robust feature representation and selection capabilities enable the model to effectively capture relevant features spanning the physical space of the vehicles. The introduction of a unidirectional topology-based causal mask is based on a reasonable assumption: during platoon movement, the vast majority of drivers base their decisions on the traffic state ahead. This assumption further strengthens the applicability of  model to platoons of varying lengths. The formal computation process of the PFL module is as follows:
\begin{equation}
    \label{eq8}
 H_{1:N,t}^{PFL} = {\varPhi  _{PFL}}(H_{1:N,t}^{FUL},{W_{PFL}})
 \end{equation}
where $H_{1:N,t}^{PFL}\in {{R}^{N \times d_{m}}}$, and denotes the output of the PFL module, with each $h_{n,t}^{PFL}$, $n = 1,2, \cdots ,N$,  encapsulating the features of vehicle $n$ and its interactions with other vehicles within the platoon.

\subsubsection{Non-Autoregressive Parameters Decoder}
The NARP decoder is tasked with decoding parameters for prediction horizon. For any following vehicle $n$ in the platoon, $h_{n,t}^{PFL}$ is duplicated along the time dimension and augmented with temporal encoding (TE). Subsequently, these vectors, containing platoon interaction features, are used as queries. Through time-based causal cross-attention, useful information from the vehicle's historical trajectory encoding $H_{n,t - P:t}^{TFL}$ is extracted for future driving decisions and projected to the dimensions corresponding to the predicted parameters via a linear layer. A time-based causal mask is necessary to prevent the model from unrealistically utilizing distant future information for prediction. The decoder employs a non-autoregressive approach for prediction to accelerate model inference, enhancing its capability for online trajectory prediction and generation. The computation process of NARP decoder resembles that shown in Fig. \ref{feature} and can be formally expressed as follows: 
\begin{equation}
    \label{eq9}
 H_{n,t:t+F-1}^{Dec} = {\varPhi  _{Dec}}(h_{n,t}^{Dec},H_{n,t - P:t}^{Dec},{W_{Dec}})\; n = 1,2, \cdots ,N
 \end{equation}
 where  $H_{n,t:t+F-1}^{Dec}\in  {{R}^{F \times3}}$ denotes the output of the decoder.

\textbf{Remark 2:} In practice, it is not always necessary to assign separate parameters for each time step in the prediction horizon. Doing so might lead the model to prioritize numerical features over dynamic platoon modeling and incur substantial computational overhead. A more reasonable approach is to share the same set of parameters across 
$m$ steps, thereby reducing the decoder's output step size to $\frac{F}{m}$.

\subsubsection{Analyzable Parameters Encoded Computational Graph}
The APeCG module begins by applying physical encoding to $H_{1:n,t:t+F-1}^{Dec}\in  {{R}^{N\times\frac{F}{m} \times3}}$. The Softplus activation function is employed to ensure these parameters satisfy the basic physical meanings and local stability conditions outlined in Eq, (\ref{eq4}). Following this, these parameters are assigned their physical meanings and used for predicting the future states of the platoon. The iterative computation process, as illustrated in Fig.  \ref{total}(d), is applied to the entire platoon. The formalized computation process is as follows: 
\begin{equation}
    \label{eq10}
\begin{array}{l}
{{\theta} _t} = [ - 1,1,1] \cdot Softplus(H_{1:n,t:t+F-1}^{Dec})\\[1ex]
a(k\left| t \right.) = {\theta _t}(k) \cdot ({x_{1:N}}(k\left| t \right.)-{x_{1:N}^* }(k\left| t \right.))\\[1ex]
\left[ \!\!{\begin{array}{*{5}{c}}
{v(k+1| t )}\\
{s(k+1| t )}
\end{array}}\!\! \right] = \left[  \!\!{\begin{array}{*{5}{c}}
1&0\\
{\Delta t}&1
\end{array}} \!\! \right] \cdot \left[  \!\!{\begin{array}{*{5}{c}}
{v(k\left| t \right.)}\\
{s(k\left| t \right.)}
\end{array}}  \!\!\right] + \left[ \!\! {\begin{array}{*{5}{c}}
{\Delta t}\\
0
\end{array}}  \!\!\right] \cdot a(k\left| t \right.)\\[2ex]
\Delta v(k+1\!\left| t\! \right.) \! = \! {[{v_0}(t+k+1\!),v_{1:N\! - \!1}(k+1\!\left| t\! \right.)^T]^T}  \!\!- \!v(k+1\!\left| t\! \right.)
\end{array}
 \end{equation}
where ${\theta _t}(k)\in {{R}^{N\times 3}}$ denotes the value of the $k$-th time step in the parameter $\theta _t$;  $k\left| t \right.$ indicates the prediction for the time $t+k$ made at time $t$.
\subsubsection{Loss Function}
The PeDL framework undergoes end-to-end training by minimizing the difference between the predicted and actual states of the platoon over the prediction horizon:  
\begin{equation}
\begin{array}{c}
    \label{eq11}
{L_{\text{Pred}}} = {\alpha _v} \cdot \frac{1}{F}\sum\limits_{k = 1}^F {\left\| {v(t + k) - v(k\left| t \right.)} \right\|_2^2} + {\alpha _s} \cdot \frac{1}{F}\sum\limits_{k = 1}^F {\left\| {s(t + k) - s(k\left| t \right.)} \right\|_2^2} 
\end{array}
 \end{equation}
where ${\left\|  \cdot  \right\|_2}$ represents the $l_2$-norm of a vector, while $\alpha _v$ and $\alpha _s$ denote the weights for the prediction losses of velocity and gap, respectively. To circumvent the challenge of selecting hyperparameters, dynamic weight averaging (DWA) \citep{liu2019end} is employed to automatically determine these values.

The variational encoder was trained using Monte Carlo sampling combined with reparameterization techniques, KL divergence is also incorporated into the loss function:
\begin{equation}
    \label{eq12}
L_{\text{KL}} = \frac{1}{2}\left( {\mu _t^2 + \sigma _t^2 - \log \left( {\sigma _t^2} \right) - 1} \right)
 \end{equation}
The loss function for PeMTFLN is defined as:
\begin{equation}
    \label{eq13}
L = {L_{\Pr {\rm{ed}}}} + {\alpha _{{\rm{KL}}}} \cdot {L_{{\rm{KL}}}}
 \end{equation}
where ${\alpha _{{\rm{KL}}}}$ is a hyperparameter that controls the degree of uncertainty in the model's output. A higher value of ${\alpha _{{\rm{KL}}}}$ promotes greater diversity in the model's output, potentially at the cost of accuracy. Conversely, a lower value enhances the model's fit but may increase the risk of overfitting.

\section{Experiments\label{prediction}} 
\subsection{Experimental design}
\subsubsection{Dataset preparation}
We extract the platoon trajectory data of seven vehicles (one leading vehicle and six following vehicles) from HIGH-SIM dataset \citep{shi2021video} for PeDL. Our extraction method follows Mo et al. \citep{mo2021physics} and adds the recognition of continuous following vehicles. As a result, the car-following scenario of each platoon in processed dataset for training and testing spans 15 seconds, with a temporal interval of 0.1 seconds. The dataset consists of 739 platoons and total trajectories encompassed more than 212751 examples.

\textbf{Remark 3:} To maximize the utilization of platoon trajectories, we conducted experiments using platoons with a uniform number of vehicles. However, the PeDL is flexible and scalable, which faces the greatest scalability constraints, the combination of attention mechanisms and tokenization techniques allows the model to effectively manage large-scale platoon dynamic modeling.

\subsubsection{Experimental setups}
The PeDL framework is evaluated with the proposed MTFLN and two classic temporal models, i.e. LSTM and Transformer. In addition, the ablation experiment of the multi-scale trajectory feature learning network (MTFLN) module has also been carried out. Note that all of models have the same dimensionality.

% \begin{itemize}
 \textbf{PeLSTM:} A neural network comprising an LSTM encoder-decoder and a variational encoder. The model encodes the trajectory of each vehicle in the platoon using the LSTM encoder, performs uncertainty learning, and subsequently autoregressively predicts the required parameters with the LSTM decoder.

\textbf{PeTransformer}: A neural network comprising a Transformer and a variational encoder, with an architecture analogous to that of PeLSTM.

 \textbf{PeMTFLN w/o TFL}: Remove TFL module from MTFLN and keep PFL module.

 \textbf{PeMTFLN w/o PFL}: Remove PFL module from MTFLN and keep TFL module.
% \end{itemize}

All models were implemented using PyTorch and trained on a server equipped with an Intel i9-13900K CPU and an NVIDIA RTX 4090 GPU. Platoon trajectory data were segmented using a sliding window approach and divided into training, validation, and test sets in a 7:1:2 ratio. The hyperparameters for PeMTFLN are detailed in Table \ref{Table I}.

\begin{table}[] 
\belowrulesep=0pt
\aboverulesep=0pt
\renewcommand\arraystretch{1.5}
\setlength{\tabcolsep}{1mm}
\caption{Hyperparameters of PeMTFLN}
\label{Table I}
\begin{tabular}{@{}>{\centering\arraybackslash}m{0.25\linewidth}|>{\centering\arraybackslash}m{0.22\linewidth}>{\centering\arraybackslash}m{0.1\linewidth}>{\centering\arraybackslash}m{0.22\linewidth}>{\centering\arraybackslash}m{0.1\linewidth}@{}}
\toprule
Module & Hyperparameter & Value & Hyperparameter & Value \\
\midrule
\multirow{4}{*}{\centering VFL Module}& Number of Mamba Block& 1& Number of SSM Sub-System& 8\\

& Number of Variational Encoder Layer & 3& Activate Function of Variational Encoder & ReLU\\

\midrule

PFL Module \& NARP Decoder & Number of Attention Layer& 2& Number of Attention Head & 4 \\
\midrule
\multirow{4}{*}{\centering Other Parameter}& Dimension of Model& 64& Learning Rate & $10^{-5}$\\
 & $m$& 5& $\alpha_{{\rm{KL}}}$& 0.0025\\

& Batch Size & 32& Max Epoch & 20\\

 & Inputs Length& 21& Outputs Length&20\\
\bottomrule
\end{tabular}
 \vspace{-1em}  % 调整与上文的间距
\end{table}

The root mean square error (RMSE) and mean absolute percentage error (MAPE) were selected as the evaluation metric:
\begin{equation}
\label{eq14}
RMSE = \sqrt {\frac{1}{N}\sum\limits_{n = 1}^N {{{\left( {{y_n} - {{\bar y}_n}} \right)}^2}} }
\end{equation}
\begin{equation}
\label{eq15}
MAPE = \frac{{100\% }}{N}\sum\limits_{n = 1}^N {\left| {\frac{{{y_i} - {{\bar y}_i}}}{{{y_i}}}} \right|}
\end{equation}
where $N$ is the number of following vehicles in platoons.

\subsection{Experimental result}
\subsubsection{Accuracy comparison}
Table \ref{metrics} presents the dynamic modeling errors metrics of different models, including RMSE and MAPE, with the best performing results highlighted in bold. PeMTFLN exhibits competitive predictive performance in both gap and velocity state prediction. Concretely, PeMTFLN achieved state-of-the-art performance in all scenarios except for speed prediction with a step size of 0.5s and 1.0s. The lowest error is observed in the PeMTFLN, records a velocity average RMSE of 0.469 m/s, velocity average MAPE of 3.09\%, gap average RMSE of 0.643 m, and gap average MAPE of 1.91\%. As the modeling step size increases, the overall trend of error accumulation increases. A noteworthy detail is that we adopt a new set of prediction parameters every 5 steps (0.5s), resulting in a velocity RMSE of 0.484m/s at 1.5s, which is a decrease compared to 0.536m/s at 1s, and this is reasonable.
Besides, we conduct ablations on PeDL. As shown in Table \ref{metrics}, TFL and PFL perform as core  modules, and discarding either of them (denoted as PeMTFLN w/o TFL and PeMTFLN w/o PFL) will result in the inferior performance. To investigate the reason, the TFL module extracts the vehicles own features in parallel from the historical trajectories of all vehicles within the platoon. The PFL module is designed to model the interaction between vehicles in the platoon and capture more macroscopic characteristics of the platoon. Therefore, PeMTFLN, with meticulously designed modules, can achieve the best performance in modeling nonlinear platoon dynamics.

\textbf{Remark 4:} Although some baseline or ablation models lack a platoon feature learning module, they can still model platoon dynamics effectively. This is because their training objective focuses on minimizing the discrepancy between the predicted and actual states of the platoon as a whole, rather than individual vehicles \citep{xu2024sequence}.
\begin{table}
\belowrulesep=0pt
\renewcommand\arraystretch{1.2}
\footnotesize
\aboverulesep=0pt
\caption{Comparison of platoon dynamic modeling accuracy indicators}\label{metrics}
\setlength{\tabcolsep}{0.2mm}{
\begin{tabular}{ c|ccccc|ccccc|ccccc|ccccc }
\toprule
\multirow{2}{*}{Model} & \multicolumn{5}{c|}{Velocity RMSE (m/s)} & \multicolumn{5}{c|}{Gap RMSE (m)}& \multicolumn{5}{c|}{Velocity MAPE (\%)} & \multicolumn{5}{c}{Gap MAPE (\%)}\\
         & 0.5s & 1.0s & 1.5s & 2.0s & Avg
         & 0.5s & 1.0s & 1.5s & 2.0s & Avg& 0.5s & 1.0s & 1.5s & 2.0s & Avg
         & 0.5s & 1.0s & 1.5s & 2.0s & Avg\\
\midrule
PeLSTM& 0.494 & 0.534 & 0.549 & 0.602 & 0.492
        & 0.396 & 0.674 & 0.943 & 1.204 & 0.689
        & 3.42 & 3.70 & 3.89 & 4.43 & 3.48
        & 1.15 & 2.03 & 2.94 & 3.88 & 2.12\\

PeTransformer& \textbf{0.483} & 0.550 & 0.507 & 0.551 & 0.479
               & 0.389 & 0.652 & 0.907 & 1.158 & 0.666
        & 3.20 & 3.67 & 3.41 & 3.86 & 3.25
        & 1.12 & 1.92 & 2.73 & 3.59 & 1.99\\
\midrule
PeMTFLN w/o TFL& 0.490 & 0.537 & 0.512 & 0.555 & 0.476
             & 0.388 & 0.651 & 0.909 & 1.162 & 0.667
        & 3.28 & 3.60 & 3.54 & 4.01 & 3.28
        & 1.12 & 1.93 & 2.78 & 3.69 & 2.02\\

PeMTFLN w/o PFL & 0.486 &\textbf{ 0.528} & 0.507 & 0.543 & 0.474
             & 0.386 & 0.651 & 0.905 & 1.154 & 0.664
        & 3.15 & 3.48 & 3.39 & 3.79 & 3.17
        & 1.11 & 1.90 & 2.72 & 3.56 & 1.98\\

PeMTFLN& 0.492 & 0.536 & \textbf{0.484} & \textbf{0.526} & \textbf{0.469}
     & \textbf{0.379} & \textbf{0.627} & \textbf{0.872} & \textbf{1.117} & \textbf{0.643}
     & \textbf{3.11} & \textbf{3.43} & \textbf{3.22} & \textbf{3.68} & \textbf{3.09}
        & \textbf{1.08} & \textbf{1.82} & \textbf{2.17}& \textbf{3.47} & \textbf{1.91} \\
\bottomrule
\end{tabular}}
  \vspace{-1em}  % 调整与上文的间距
\end{table}

% \begin{table*}[h]
% \renewcommand\arraystretch{1.5}
% \caption{HDVs driving state prediction metrics comparison}\label{Table IV}
%   \centering
% \setlength{\tabcolsep}{1.5mm}{
% \begin{tabular}{@{} c|ccccc|ccccc@{} }
% \toprule
% \multirow{2}{*}{Model} & \multicolumn{5}{c|}{Velocity RMSE (m/s)} & \multicolumn{5}{c}{Gap RMSE (m)}\\
%          & 0.5s & 1.0s & 1.5s & 2.0s & Average
%          & 0.5s & 1.0s & 1.5s & 2.0s & Average\\
% \midrule

% PeDL & 0.492 & 0.536 & 0.484 & 0.526 & \textbf{0.469}
%      & 0.379 & 0.627 & 0.872 & 1.117 & 0.643\\
% PeDL-only & \textbf{0.415} & \textbf{0.476} & 0.532 & 0.599 & 0.470
%              & 0.387 & 0.662 & 0.906 & 1.124 & 0.664\\

% PeDL-every & 0.512 & 0.525 & \textbf{0.471} & \textbf{0.516} & 0.478
%      & \textbf{0.375} & \textbf{0.615} & \textbf{0.854} & \textbf{1.099} & \textbf{0.632}\\
% \bottomrule
% \end{tabular}}
% \end{table*}
\begin{figure}[]
  \centering
  \includegraphics[width=0.8\textwidth]{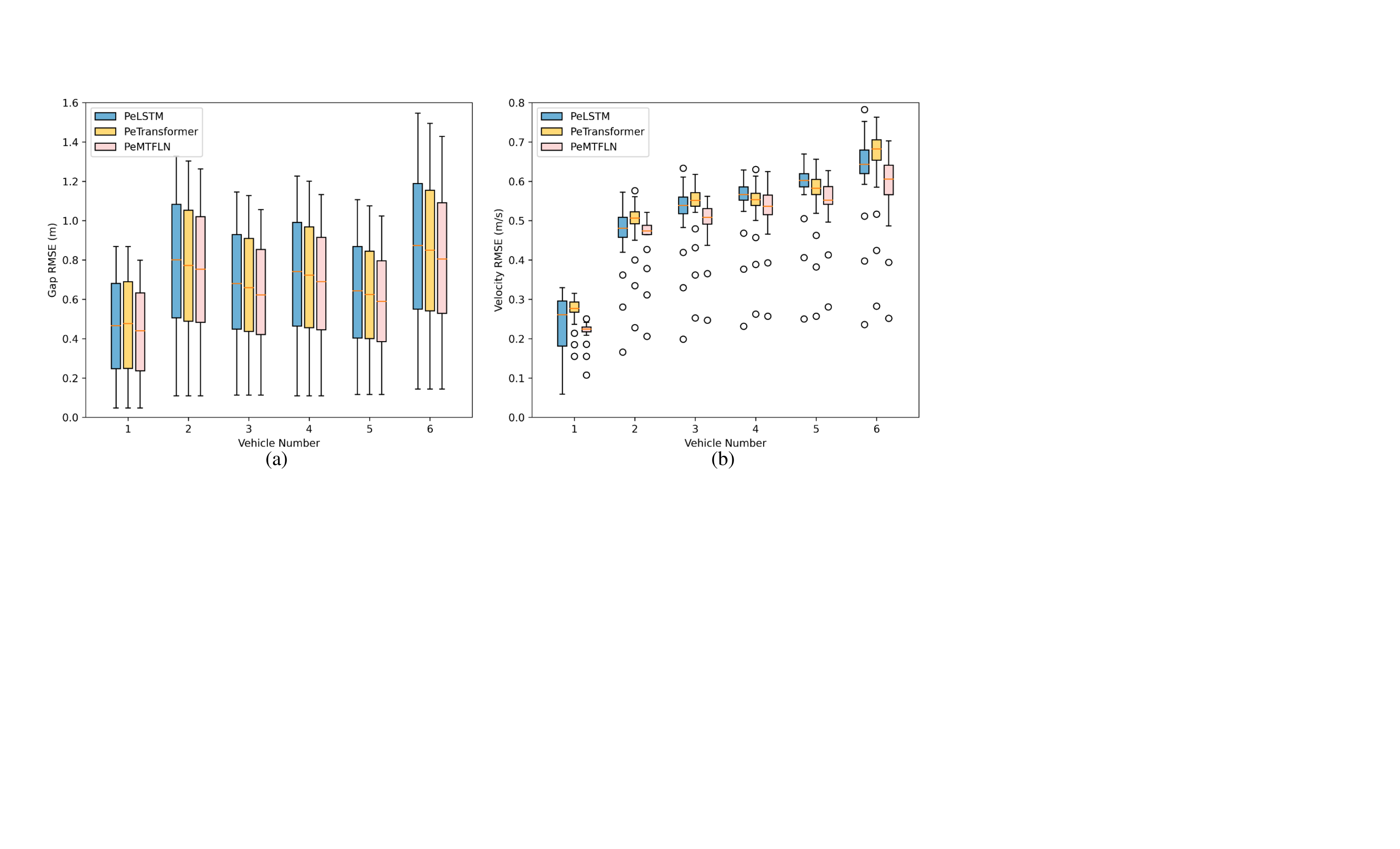}
  \vspace{-1em}  % 调整与上文的间距
  \caption{Platoon modeling errors among PeMTFLN, PeLSTM and PeTransformer. (a) Gap; (b) Velocity}\label{box}
  \vspace{-0.5em}  % 调整与上文的间距
\end{figure}
Fig. \ref{box} is a multivariate box plot used to illustrate the platoon dynamic modeling errors under PeMTFLN, PeLSTM and PeTransformer models. The horizontal axis represents the vehicle number in the platoon, and the vertical axis represents the error index. Each box contains 20 indicator values, corresponding to the modeling error for each time step from step 1 to step 20. Firstly, regarding the gap modeling in Fig. \ref{box} (a), it is evident that the proposed PeMTFLN consistently maintains accuracy over whole platoon, as reflected by the lowest median error. Secondly, as shown in Fig. \ref{box} (b), the errors have gradually accumulated in speed as the vehicle number increases. The whiskers of PeLSTM are relatively large, which means that the error values of each step are the most discrete, and the error increases rapidly along the step size. Surprisingly, PeMTFLN has the lowest median, emphasizing the superior ability of PeMTFLN in learning nonlinear platoon dynamics evolution. Overall, PeMTFLN excels at capturing the short-term and long-term behavioral characteristics, achieving precise platoon-level car-following modeling.
\begin{figure}[]
  \centering
  \vspace{-1em}  % 调整与上文的间距
  \includegraphics[width=0.8\textwidth]{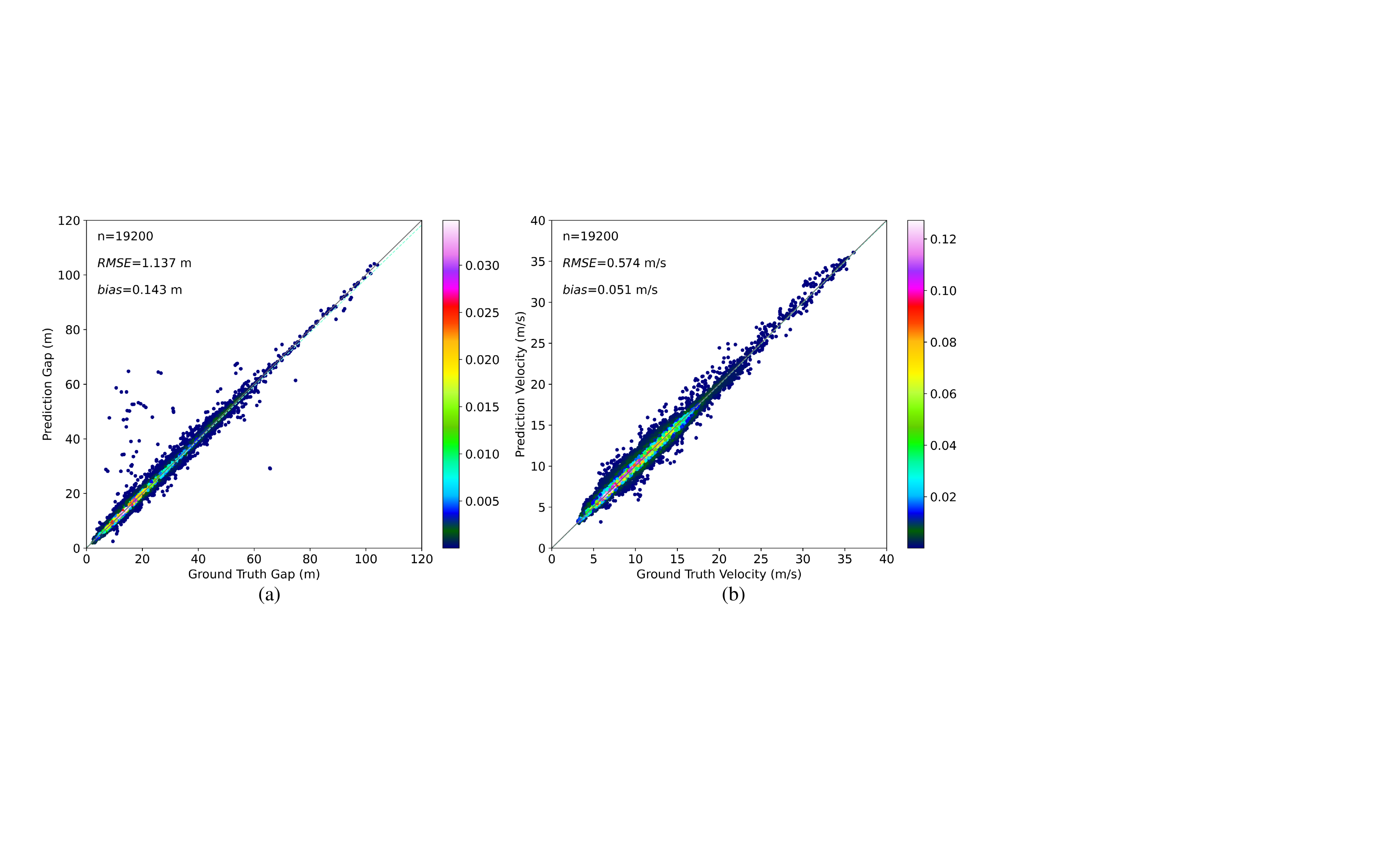}
  \vspace{-1em}  % 调整与上文的间距
  \caption{Visualization of platoon trajectory samples replicated by PeMTFLN and real trajectory samples. (a) Gap; (b) Velocity}\label{wholevis}
  \vspace{-0.5em}  % 调整与上文的间距
\end{figure}
Without loss of generality, Fig. \ref{wholevis} randomly selected 100 batches of samples (a total of 100X32X6=19200) for visual comparison of ground truth and predicted states, with the aim of demonstrating the overall accuracy of PeMTFLN for platoon dynamic modeling. Note that the corresponding modeling step size of Fig. \ref{wholevis}  is 2s. The overall observation is that the samples are clustered on the diagonal, with only a very small number of samples scattered around the diagonal, whether it is gap samples or velocity samples. The predicted results of PeMTFLN are highly consistent with the ground truth of the platoon. Quantitatively, the RMSE for gap and speed are 1.137m and 0.574m/s, respectively, with corresponding bias of 0.143m and 0.051m/s. Generally, the deviations of 0.143m and 0.051m/s are acceptable in applications such as control or simulation testing. Therefore, it can be confirmed that the proposed PeMTFLN can reproducing real driver responses with high accuracy at the platoon-level.

\begin{figure}[]
  \centering
  \includegraphics[width=0.8\textwidth]{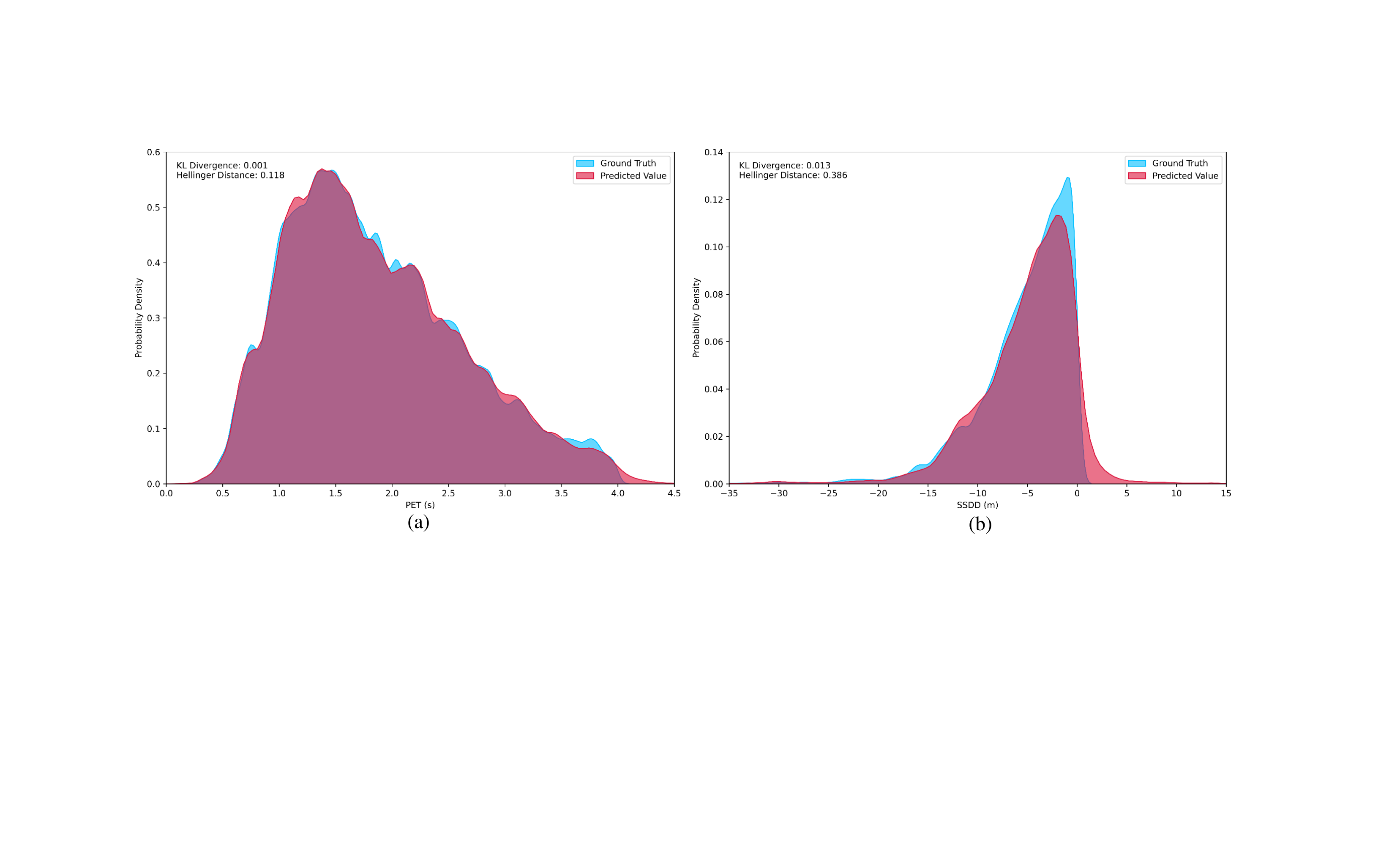}
  \vspace{-1em}  % 调整与上文的间距
  \caption{The safety statistics of platoon evolution replicated by PeMTFLN. (a)PET; (b) SSDD}\label{safety}
\end{figure}
\subsubsection{The statistical results of surrogate safety measurements}
As shown in Fig. \ref{safety}, the performance of the proposed PeMTFLN in predicting safety-critical events is validated, particularly in near-miss situations. Two surrogate safety measurements, post-encroachment time (PET) \citep{allen1978analysis} and safe stopping distance difference (SSDD) \citep{mahmud2017application}, are examined to characterize conflict-based vehicle safety \citep{guo2019modeling}. In addition, Hellinger distance and KL-divergence are used to quantitatively compare the safety statistics between predicted results and empirical ground truth data. The smaller the Hellinger distance and KL-divergence, the more similar the distributions are. It was found that the predicted PET and SSDD distribution can accurately reproduce dangerous driving conditions in actual scenarios. Quantitative comparison shows that KL divergence is 0.001 and Hellinger distance is 0.118 for the PET measure. In terms of SSDD measure, KL divergence is 0.013 and Hellinger distance is 0.386. The results validate the accurate modeling of the proposed PeMTFLN in terms of platoon safety.

\subsection{Visualization and Interpretability Analysis}
%可视化
\subsubsection{Sample Visualization}
\begin{figure}
  \centering
    \vspace{-1em}  % 调整与上文的间距
  \includegraphics[width=1\textwidth]{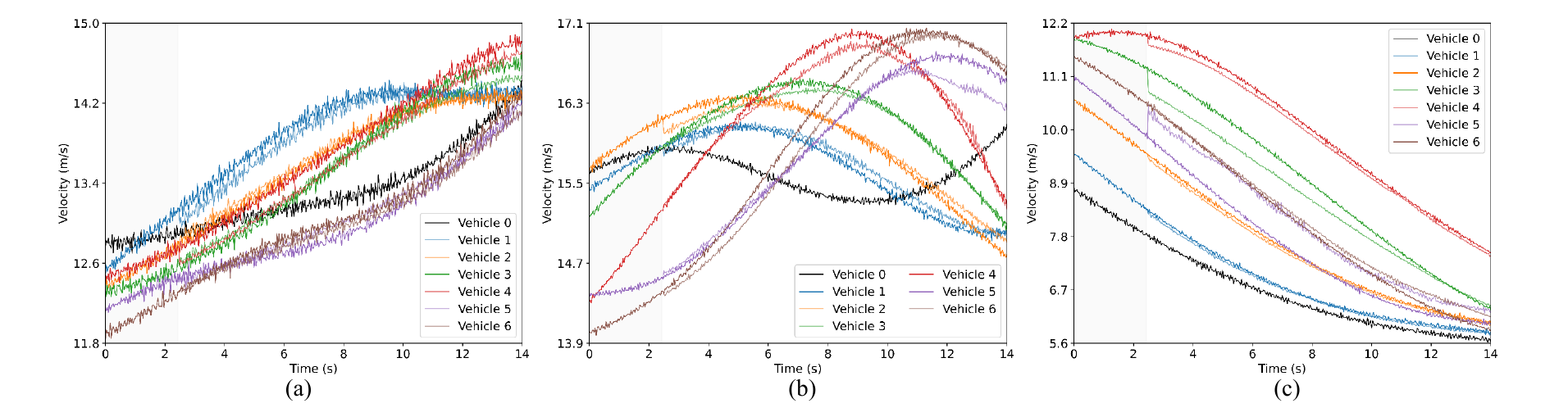}
  \vspace{-2em}  % 调整与上文的间距
  \caption{The reproduced platoon trajectory examples of PeMTFLN. (a) continuous acceleration; (b) oscillation; (c) continuous deceleration}\label{localvis}
  \vspace{-1.5em}  % 调整与上文的间距
\end{figure}

Fig. \ref{localvis} selected three platoons for comparison between real trajectories and predicted trajectories by PeMTFLN. Each sub-figure involves a different platoon driving pattern: continuous acceleration, oscillation, and continuous deceleration. The same color scheme with different shades represents the real and predicted trajectories of a certain vehicle. It is evident that PeMTFLN can maintain high accuracy for a considerable period of time. The trajectory generated by PeMTFLN (light color) is very close to the real ground situation (dark color), demonstrating that PeMTFLN can respond positively to changes in the speed of the leading vehicle within the platoon. The local visualization results for each vehicle in the platoon are consistent with the quantitative results presented in Table \ref{metrics} and Fig. \ref{wholevis}, with minimal modeling deviation for speed and gap. This is mainly attributed to the effective capture of relevant features across the physical space of vehicles by PeMTFLN, ultimately achieving effective and accurate modeling of nonliear platoon dynamics.
\begin{figure}
  \centering
  \includegraphics[width=0.95\textwidth]{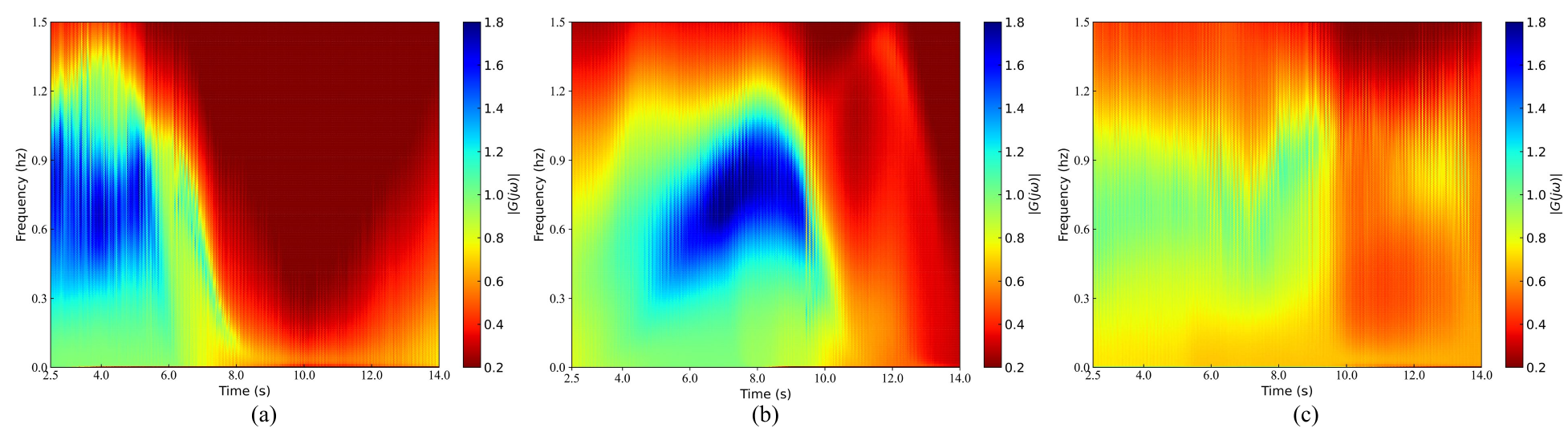}

  \caption{The platoon stability evolution replicated by PeMTFLN. (a) continuous accel- eration; (b) oscillation; (c) continuous deceleration}\label{stability}
  \vspace{-1.5em}  % 调整与上文的间距
\end{figure}
\subsubsection{Evidence for physically analyzable: stability analysis}
The string stability analysis is closely related to traffic flow modeling, as the stop and go oscillations in actual traffic are a manifestation of the traffic flow instability. From the perspective of developing a realistic platoon-level car-following dynamics model, it should have the ability to describe the string instability and replicate characteristics related to traffic oscillations. As such ability of PeMTFLN to perform stability analysis. Specifically, the transfer function magnitude value $\left| {G\left( {jw} \right)} \right|$ is calculated, which describe the evolution of disturbance transmission from the head vehicle to the tail vehicle \citep{montanino2021string}. The results are depicted in Fig. \ref{stability}. This corresponds to Fig. \ref{localvis}, Fig. \ref{stability} includes three platoon evolution scenarios: continuous acceleration, oscillation, and continuous deceleration. $\left| {G\left( {jw} \right)} \right|$ greater than 1 (the blue area) indicates that the disturbance has been amplified, reflecting the characteristics of string instability. Clearly, as shown in Fig. \ref{stability} (b), the platoon exhibits string instability, where disturbances propagate and amplify through the vehicle strings due to frequently traffic oscillations. Additionally, the vehicle platoon in the deceleration scenario has converged to a relatively consistent speed. Another finding is that the stability changes over a continuous period are not very significant. Overall, the accurate reproduction of stability in real vehicle platoon evolution scenarios emphasizes the theoretically analyzable ability of the proposed PeMTFLN.

\section{Simulation\label{Simulation}}
\subsection{Simulation setting}
Another expectation for PeMTFLN is that it can generate high fidelity trajectories in simulation environments. Thus, there is a pressing need to further evaluate the capability of PeMTFLN in "platoon trajectories generation" \citep{lin2020platoon}, which has great differences with trajectory prediction (Section \ref{prediction}). Specifically, platoon simulation refers to providing only the initial states of the leading vehicle, while following vehicles within the platoon are simulated by their initial states (2 seconds here) and the simulated results of the vehicle ahead. Notably, the baseline selected is the online-calibrated IDM. In particular, an online genetic algorithm (GA)-based personalized calibration procedure is designed to determine the IDM parameters for each vehicle (the trajectory length used is consistent with PeMTFLN), aiming to maximize the restoration of personalized driving behavior. Without loss of generality, three platoon simulation cases are randomly selected, each involving a different lead vehicle speed evolution pattern: continuous acceleration (Platoon 8), deceleration (Platoon 10) and oscillation (Platoon 14). 
\begin{figure}
  \centering
  \vspace{-2em}  % 调整与上文的间距
  \includegraphics[width=1\textwidth]{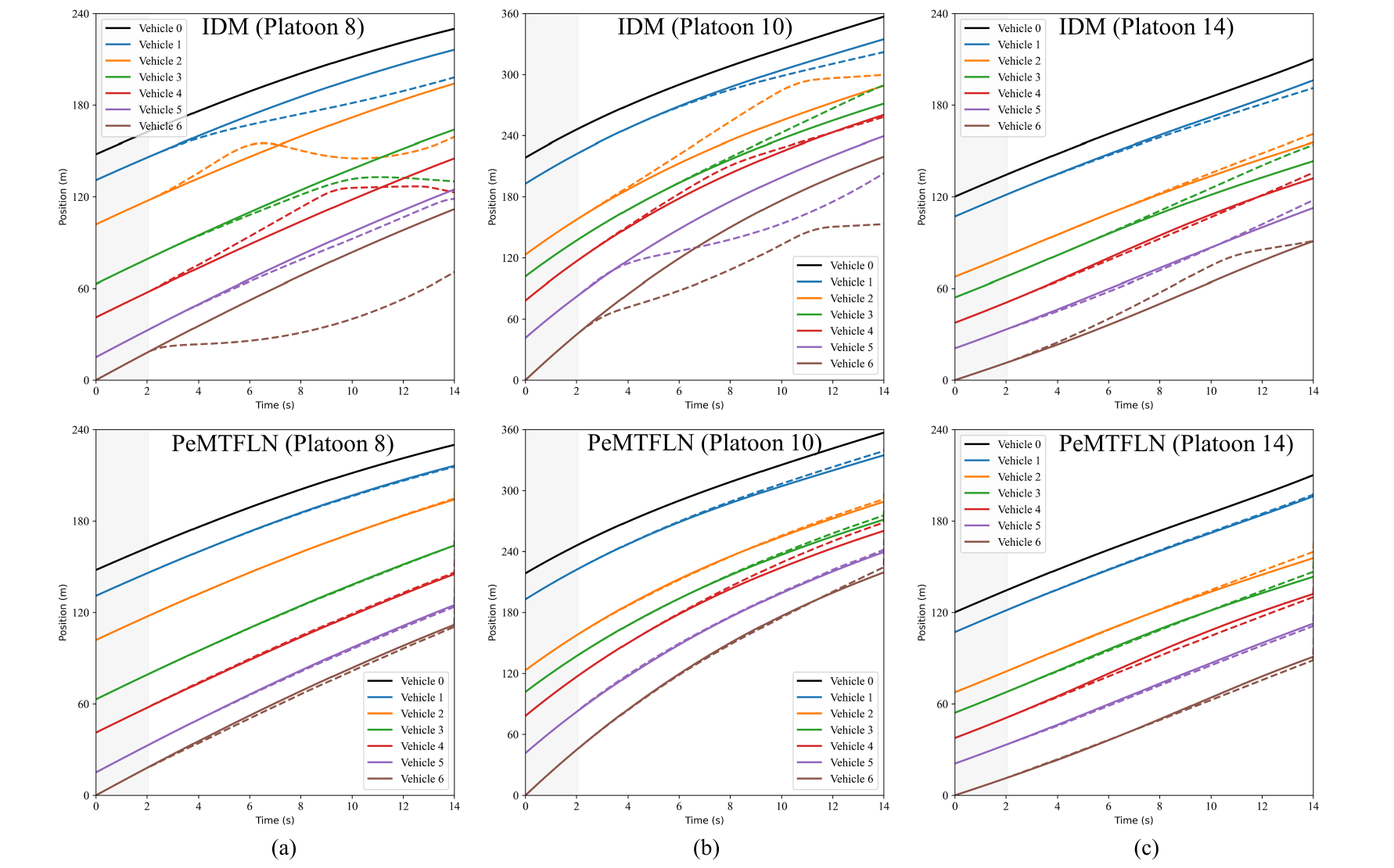}
    \vspace{-2em}  % 调整与上文的间距
  \caption{The reproduced position evolution of IDM and PeMTFLN. (a) continuous acceleration; (b) oscillation; (c) continuous deceleration }\label{generation_position}
      \vspace{-0.5em}  % 调整与上文的间距
\end{figure}
\begin{figure}
  \centering
   \vspace{-1em}  % 调整与上文的间距
  \includegraphics[width=1\textwidth]{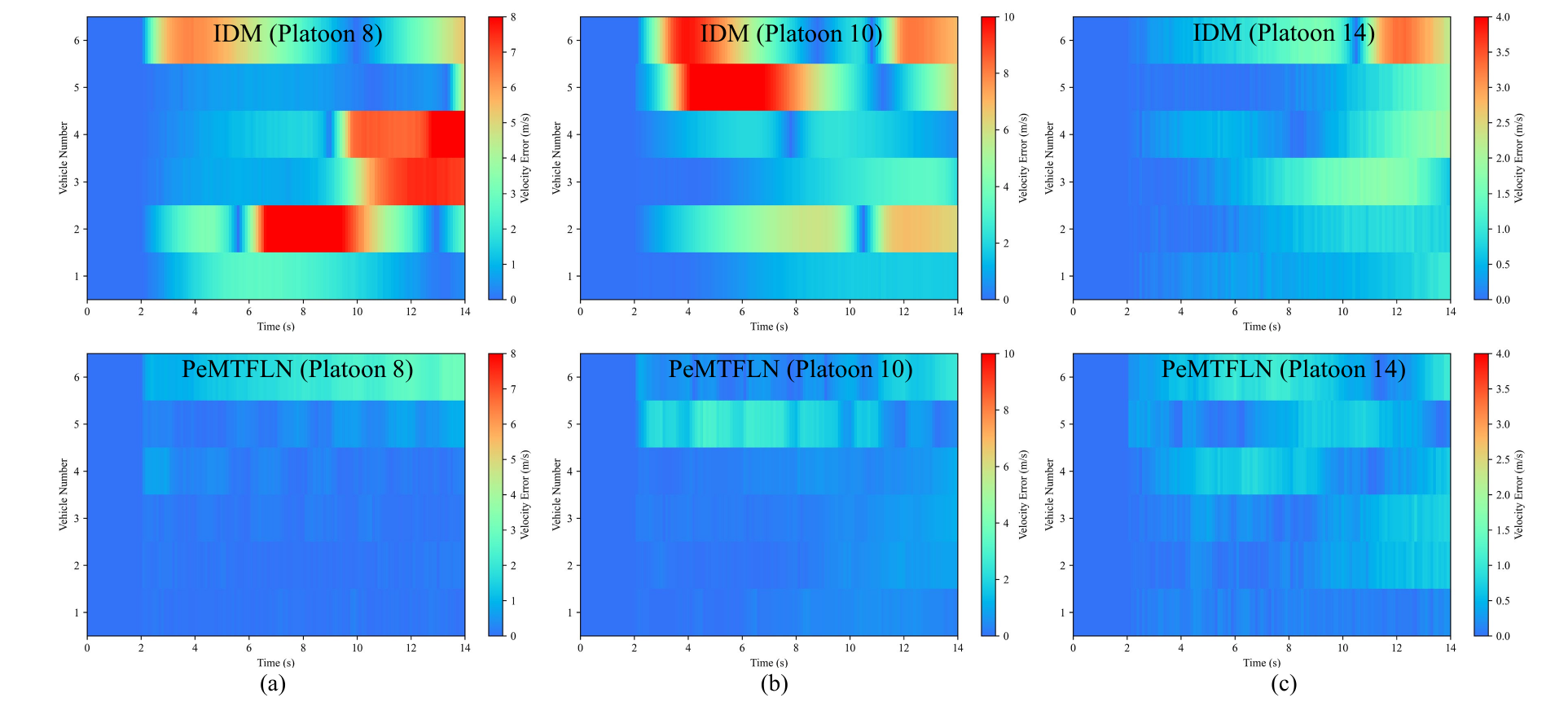}
    \vspace{-2em}  % 调整与上文的间距
  \caption{The reproduced speed evolution of IDM and PeMTFLN. (a) Continuous acceleration; (b) Oscillation; (c) Continuous deceleration}\label{generation_speed}
   \vspace{-1.5em}  % 调整与上文的间距
\end{figure}
\subsection{Simulation results}
Fig. \ref{generation_position} presents the simulation results of three scenarios where the solid lines are the real trajectories while the dashed lines are the simulated trajectories. The contrast of PeMTFLN control strategy and IDM control strategy is strongly intuitive. With the IDM control, the evolution of the platoon exhibits overly conservative or overly aggressive driving characteristics, leading to deviations from the real trajectories. On the contrary, the generated trajectories by PeMTFLN close follow to the real trajectories with remarkable precision. In addition, it was found that for scenarios where the speed of the leading vehicle varies greatly, as shown in Fig. \ref{generation_position} (a) and Fig. \ref{generation_position} (b), IDM is difficult to accurately simulate the trajectory. However, in Fig. \ref{generation_position} (c), for scenarios where the speed oscillates within a certain range, IDM can regenerate the actual platoon trajectory. Moreover, with PeMTFLN control, the generated trajectories of the platoon initially align so perfectly with the real trajectories that the dashed lines are completely overlapped with the solid lines. As time progresses, the dashed lines deviates slightly from the solid lines, which is make sense and corresponds to the actual situation as errors tend to increase with longer time steps. In summary, PeMTFLN overcomes the limitations presented by IDM, with precise inference in trajectory generation and the ability to reproduce complex traffic phenomena. PeMTFLN solves the prevailing error propagation phenomenon in both time and space dimensions, thereby enhancing the accuracy and reliability of practical simulation applications. 
\begin{figure}
  \centering
  \vspace{-2em}  % 调整与上文的间距
  \includegraphics[width=1\textwidth]{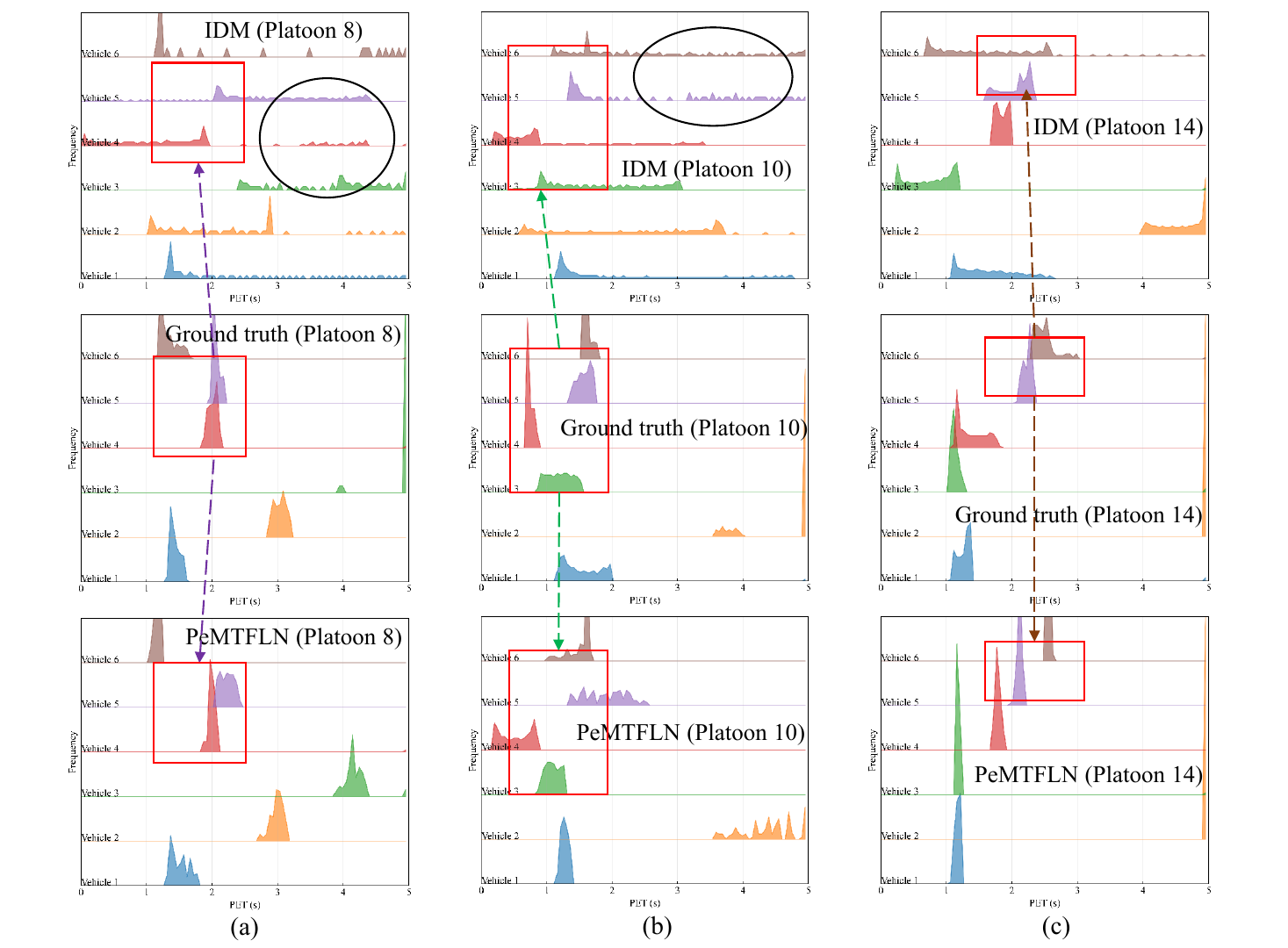}
 \vspace{-2em}  % 调整与上文的间距
  \caption{The PET distributions of reproduced  trajectory with IDM and PeMTFLN. (a) Continuous acceleration (b) Oscillation (c) Continuous deceleration}\label{generation_pet}
  \vspace{-1.5em}  % 调整与上文的间距
\end{figure}
Fig. \ref{generation_speed} is presented from a platoon speed generation perspective, corresponds to Fig. \ref{generation_position}, which depicts the deviation value between the speed simulated by IDM  and PeMTFLN and the actual trajectory. The color bar represents the gradient of simulation error, with blue indicating minimal error and red denoting maximum error. In Fig. \ref{generation_speed} (a) and Fig. \ref{generation_speed} (b), significant discrepancies exist between the simulated platoon evolution results under the IDM control strategy and the ground truth in the randomly selected platoon 8 case and platoon 10 case, particularly as time progresses and errors accumulate. IDM primarily focuses on the acceleration and deceleration behavior of individual vehicles, neglecting the overall coordination of the platoon. Of course, in Fig. \ref{generation_speed} (c), the lead vehicle exhibits small amplitude speed oscillations, IDM also achieves relatively satisfactory control performance, but the propagation of simulation errors is evident. As expected, PeMTFLN achieved accurate simulations across various cases, owing to its effectiveness in extracting both vehicle-level and platoon-level interaction features from natural driving platoon trajectories. The high-precision simulation of platoon dynamic by PeMTFLN provides opportunities for potential subsequent microscopic simulation applications.

Fig. \ref{generation_pet} depicts the post-encroachment time (PET) distributions of simulated trajectory by IDM and PeMTFLN, as well as the PET distribution for ground truth(middle). The PET is a widely used surrogate safety measure for identifying near-miss situations. As highlighted by the red box, the PET distribution of the vehicle platoon trajectories simulated by PeMTFLN closely matches the actual distribution. The distribution deviations in IDM simulation, marked by the black boxes, are to some extent due to the poor adaptability of IDM to external disturbances and lack of support for multi-vehicle coordinated control. From a broader perspective, PeMTFLN can characterize real-world near-accident statistics, which validates the modeling accuracy of the proposed PeMTFLN regarding platoon safety-critical behaviors.

\section{Conclusion\label{Conclusion}}
This paper proposes a novel framework that encodes physics analyzable parameters into deep learning networks for modeling the nonlinear platoon-level dynamics. Specifically, we parameterize a physically interpretable generalized platoon dynamic model and incorporate it into the computational graph (APeCG). The MTFLN is designed to learn the parameters required for APeCG from naturalistic platoon driving trajectories, capturing dynamic features from both vehicle and platoon-level perspectives. Additionally, MTFLN uses an efficient non-autoregressive decoder for multi-step prediction. The HIGHSIM dataset is employed for the end-to-end training of PeMTFLN. Experimental results demonstrate that PeMTFLN consistently outperforms baseline models, achieving a maximum average gap error of only 1.117 m and a maximum average velocity error of 0.526 m/s over a 2s horizon. The analysis of distance-based and time-based surrogate safety measures showed that the reproduced distributions were very closely aligned, with KL divergences of only 0.013 and 0.001, respectively. Furthermore, for platoon samples with different driving modes, PeMTFLN not only accurately replicates the raw trajectories of the platoon but also provides consistent head-to-tail stability analysis results of platoon with observed results. Moreover, simulation results also demonstrate that the proposed PeMTFLN exhibits extremely small inference error in platoon trajectory generation, outperforming the capabilities of IDM.

Despite its capabilities, the current PeMTFLN model does have some limitations. For instance,  APeCG is grounded in linear vehicle dynamics and requires iterative trajectory generation. Prospectively, we plan to redefine APeCG utilizing more precise vehicle dynamics and selective scanning algorithms. Another promising idea is integrating PeMTFLN with scenario generation techniques to produce platoons characterized by specific driving styles or safety-critical scenarios. This is expected to become an effective tool for the development and testing of control algorithms for autonomous vehicle platoon.
 
\section*{Acknowledgement}
This research was supported by the project of 
the National Key R\&D Program of China (No. 2023YFB4302701), the National Natural Science Foundation of China (No. 51925801, 52232012, 52272343, 52131203), SEU Innovation Capability Enhancement Plan for Doctoral Students (No.CXJH\_SEU 25178) and Postgraduate Research \& Practice Innovation Program of Jiangsu Province (No. KYCX24\_0452) .

\bibliographystyle{cas-model2-names}

\bibliography{main}

\end{sloppypar}
\end{document}